\documentclass[3p,times,procedia]{elsarticle}
\usepackage{booktabs}
\usepackage{ecrc}
\usepackage[bookmarks=false]{hyperref}
    \hypersetup{colorlinks,
      linkcolor=blue,
      citecolor=blue,
      urlcolor=blue}

\volume{00}

\firstpage{1}

\journalname{Procedia Computer Science}

\runauth{Zhe Zhang}

\jid{procs}

\usepackage{amssymb}

\usepackage[figuresright]{rotating}

\begin{document}
\begin{frontmatter}



\dochead{Proceedings of International Conference on Biomimetic Intelligence and Robotics}

\title{Mani-GPT: A Generative Model for Interactive Robotic Manipulation}


\author[a,b]{Zhe Zhang} 
\author[d,e]{Wei Chai}
\author[a,b,c]{Jiankun Wang\corref{cor1}}

\address[a]{Shenzhen Key Laboratory of Robotics Perception and Intelligence, Shenzhen, China}
\address[b]{Department of Electronic and Electrical Engineering, Southern University of Science and Technology, Shenzhen, China}
\address[c]{Jiaxing Research Institute, Southern University of Science and Technology, Jiaxing, China}
\address[d]{Senior Department of Orthopedics, the Fourth Medical Center of PLA General Hospital, Beijing, China}
\address[e]{National Clinical Research Center for Orthopedics, Sports Medicine and Rehabilitation, Beijing, China}

\begin{abstract}
In real-world scenarios, human dialogues are multi-round and diverse. Furthermore, human instructions can be unclear and human responses are unrestricted. Interactive robots face difficulties in understanding human intents and generating suitable strategies for assisting individuals through manipulation. In this article, we propose Mani-GPT, a Generative Pre-trained Transformer (GPT) for interactive robotic manipulation. The proposed model has the ability to understand the environment through object information, understand human intent through dialogues, generate natural language responses to human input, and generate appropriate manipulation plans to assist the human. This makes the human-robot interaction more natural and humanized. In our experiment, Mani-GPT outperforms existing algorithms with an accuracy of 84.6\% in intent recognition and decision-making for actions. Furthermore, it demonstrates satisfying performance in real-world dialogue tests with users, achieving an average response accuracy of 70\%.
\end{abstract}

\begin{keyword}
ambiguous intent recognition, interactive manipulation, generative model; 




\end{keyword}
\cortext[cor1]{Corresponding author.}
\end{frontmatter}

\email{wangjk@sustech.edu.cn}



\section{Introduction}
\label{main}

Ideally, interactive robots can dialogue naturally with humans, recognize human intents and perform proper manipulation to complete the tasks. However, human dialogues are often multi-round, usually involving unclear instructions and free-form responses \cite{he2019hybrid, jiang2019smart} . Most existing interactive robots can only handle single-round dialogues and fixed question-answering format. This makes the dialogues between humans and robots unnatural and limits the applications of these robots.

Many studies have focused on deploying an interactive module on robot manipulation systems. \cite{zellers2021piglet, huang2022language, shridhar2023perceiver, zhao2023chat, min2021film, min2022don} . The goal is to enhance robots' ability to understand the environment and improve their planning capabilities for manipulation tasks. Shridhar et al. \cite{shridhar2018interactive, shridhar2020ingress} propose a robotic arm grasping system based on natural language interaction, named INGRESS. This system infers objects and their relationships from input images and linguistic instructions, allowing it to recognize ambiguous human intents and perform grasping tasks without constraints. Zhang et al. \cite{zhang2021invigorate} propose INVIGORATE, a robotic system that can ask questions to better identify and grasp the target object. To determine whether to ask or grasp, this system utilizes a Partially Observable Markov Decision Processes (POMDP). This decision-making approach allows it to consider the uncertainty in its observations and make informed choices based on the available information. Yang et al. \cite{yang2022interactive} propose Attr-POMDP, also allowing the model to ask questions about poorly expressed tasks and to capture what humans want. However, these three tasks are limited to comprehending basic human descriptions, such as appearance, in order to identify relevant objects. They have limited ability to understand ambiguous and complex human instructions. Mo et al. \cite{mo2023towards} propose SeeAsk to deal with open-world scenarios. It is also based on POMDP, but they can solve more complex tasks. For example, for ambiguous object information, they can ask questions to complete the information in terms of spatial location relationships, appearance, color, etc. However, SeeAsk uses an interaction method that uses Markov decisions to select questions to ask from a fixed pool of questions, while accepting a relatively fixed response format. This makes it unable to handle human colloquial expressions, jumpy topics and multi-round dialogues in natural dialogues.

In this work, we propose \textbf{Mani-GPT}: \textbf{Mani}pulation \textbf{G}enerative \textbf{P}re-\textbf{T}rained Model, a generative model for natural language dialogue and interactive manipulation. Mani-GPT is inspired by the GPT model for language generation, but it is specifically trained to generate both dialogue responses and manipulation plans to according to human intents. Mani-GPT is capable of understanding human intents by engaging in multi-round dialogues. It can formulate strategies to assist humans by selecting appropriate actions based on the dialogue context. Additionally, it is trained to generate human-like responses, which helps create more natural and engaging dialogues that are similar to those between humans. In the generation process, Mani-GPT utilizes an object detection model to understand the environment. It then chooses proper actions from a range of available options, such as grasping objects, answering questions, and providing helpful responses. With this ability, Mani-GPT can handle ambiguous and complex dialogues, accurately interpret ambiguous human intentions, and provide more thoughtful manipulation plan to assist humans.

The contributions of this article are as follows:
\begin{itemize}

\item An LLM-based interactive model for human intent understanding and manipulation plan generating is proposed. It works by comprehending ambiguous intentions through dialogues, producing natural language responses to instructions, and using appropriate manipulation actions to assist humans.

\item A high quality dialogue dataset is designed. The dataset contains 20k single-turn and multi-round dialogue data between humans and Artificial Intelligence (AI) assistant, which is used to train the model.

\item A pipeline is built to realize an interactive manipulation process. The pipeline consists of a visual understanding module, an action decision module and a response generation module. Through this pipeline, Mani-GPT understands human intent through dialogues, and executes proper actions like grasping and answering to assist human.
\end{itemize}

\section{Method}
The overview of Mani-GPT is shown in Fig. 1. The pipeline includes a visual module, an action decision module and a response generation module. At each time step $t$, the visual module returns labels and bounding boxes. Action decision module takes the labels $l_t^i, i = 1, 2, {\cdots}, N$ from the visual module outputs and takes language information $o_t$ from the dialogue history information, then generates manipulation actions $a_t^j, j = 1, 2, {\cdots}, N$. The response generation module executes the actions, then takes $l_t^i$, $o_t$ and $a_t^j$ to generate corresponding responses $r_t^k, k = 1, 2, {\cdots}, N$. 

 In this setup, we build up the pipeline with the following works: (1) An object detection model is used for Mani-GPT to visually understand the environment. Visual information is transmitted to both the action decision module and the response generation module. (2) Mani-GPT is deployed to determine appropriate actions and generate responses. By modifying the output of the GPT model, Mani-GPT plays a role in both action decision module and response generation module. The action decision module generates suitable manipulation actions, and the response generation module executes these actions and generates corresponding responses. (3) A set of executable manipulation actions are designed to simulate the assist process of a service robot in a human society, including grasping, responding, confirming and refusing. (4) A dataset is created to train the action decision module. Each of the works is described in detail below. 

\subsection{Visual Understanding}

The visual module employs an object detection model (such as Detic \cite{zhou2022detecting} ) to identify objects in the surroundings and provide object labels. Labels are provided to the action decision module to help it develop a semantic understanding of the environment and obtain useful knowledge from language. To provide guidance for the Mani-GPT model in understanding the environment, object labels are included in the model's prompt. For example, a prompt can be written as: ``You are in a kitchen. You can see an apple, a knife and ... on the table". The prompt with object labels can affect the distribution of the generative model, leading to responses that align with the object labels.

Bounding boxes are given to the response generation module to provide spatial information for grasping. When Mani-GPT intends to perform a grasping action targeting specific objects, the grasping model will detect those objects within their respective bounding boxes. This process significantly aids the performance of the grasping network.

\begin{figure}[t]\vspace*{4pt}
\label{fig1}
\centerline{\includegraphics[width=\textwidth]{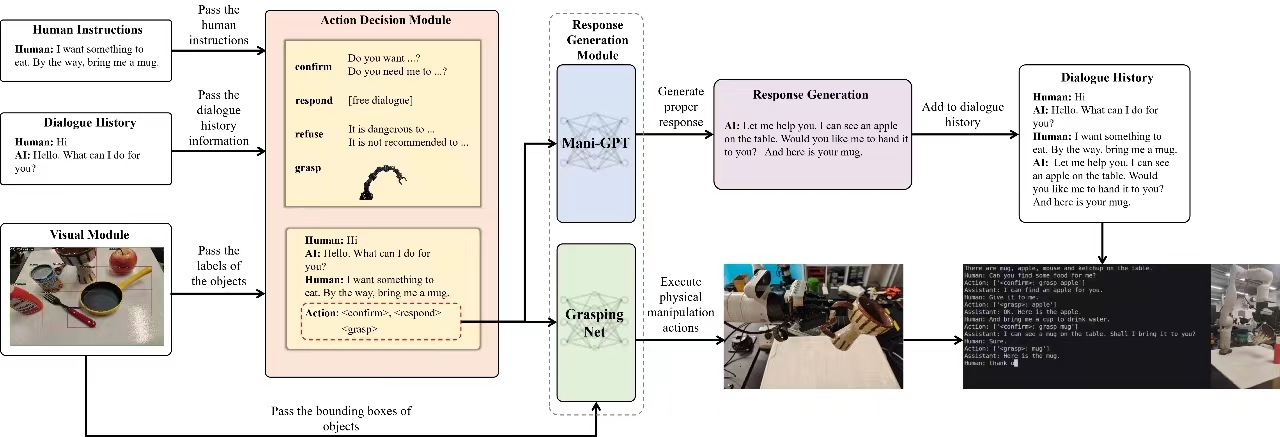}}
\caption{Mani-GPT workflow. The object detection labels from the visual module and dialogue history are combined and sent to the action decision module. Using this input, the action decision module chooses the most appropriate manipulation actions, while the response generation module generates a response based on the input and executes the actions. The generated responses are then added to the dialogue history.}
\end{figure}
\subsection{Action Decision and Response Generation}  

Mani-GPT is responsible for determining actions to interact with the environment and generating responses during dialogues with humans. To generate actions and responses that are more relevant to the environments and the tasks, we make changes to the way the generative model outputs its results. For our Mani-GPT, the probability of the next token is changed as:

\begin{equation}
\label{equ}
p(\mathcal{O}) = \prod_{i=1}^{n} P(r_t \mid \ p, \{h_i\}_{i=1}^{n-1}, q_{n}), 
\\(h_i = q_i, r_i, h_0 = \varnothing),
\end{equation} 

where $r$ represents the output of the model. 
$h = \{h_i\}_{i=1}^{n}$represents the dialogue history, $p$ represents the constructed inquiry template that encompasses available object labels and $q_n$ represents the human's query or request. 

The modified generative model can incorporate environmental information and manipulation actions through dialogues and prompts. However, it is possible that the generated responses may include actions or objects that do not exist. To limit the output of the model, we split the prediction process into two parts: generating manipulation actions based on the humans' query, and generating responses using both the query and the actions. This approach simplifies the training process and enables the model to converge more easily. The probability of the response is changed to:

\begin{equation}
\label{equ}
p(\mathcal{R}) = \prod_{k=1}^{n} P(r_t^{k}  \mid p, \{h'_i\}_{i=1}^{t-1}, q_t, a_t),
\\(h_i = q_i, a_i, r_i, h_0 = \varnothing),
\end{equation}

where $r_t^{k}$ represents the $k$-th word of the response at time step t. And the probability of the actions can be represented as:
\begin{equation}
\label{equ}
p(\mathcal{A}) = \prod_{j=1}^{n} P(a_t^j  \mid p, \{h'_i\}_{i=1}^{t-1}, q_t),
\\(h_i = q_i, a_i, r_i, h_0 = \varnothing),
\end{equation}

where $a_t^j$ represents the $j$-th action of the actions at time step $t$.

In order to improve the accuracy of the model in predicting manipulation action sequences and responses simultaneously, we design the loss function as:

\begin{equation}
\label{equ}
L = \sum_{i=1}^{t} \log P(a_t  \mid p, \{h'_i\}_{i=1}^{t-1}, q_t; \mathrm{\Theta_1})) +\lambda \sum_{i=1}^{t} \log P(r_t  \mid p, \{h'_i\}_{i=1}^{t-1}, q_t, a_t; \mathrm{\Theta_2})
) = L_a + \lambda L_r,
\end{equation}

where $\lambda$ is a constant to balance the two loss functions. The use of a specialized loss function can enhance the model's reasoning capabilities and enable it to integrate its own executable actions, leading to more effective problem-solving and the generation of suitable response utterances. This helps the model develop a robust reasoning process.

\begin{figure}[t]\vspace*{4pt}
\label{fig2}
\centerline{\includegraphics[width=\textwidth]{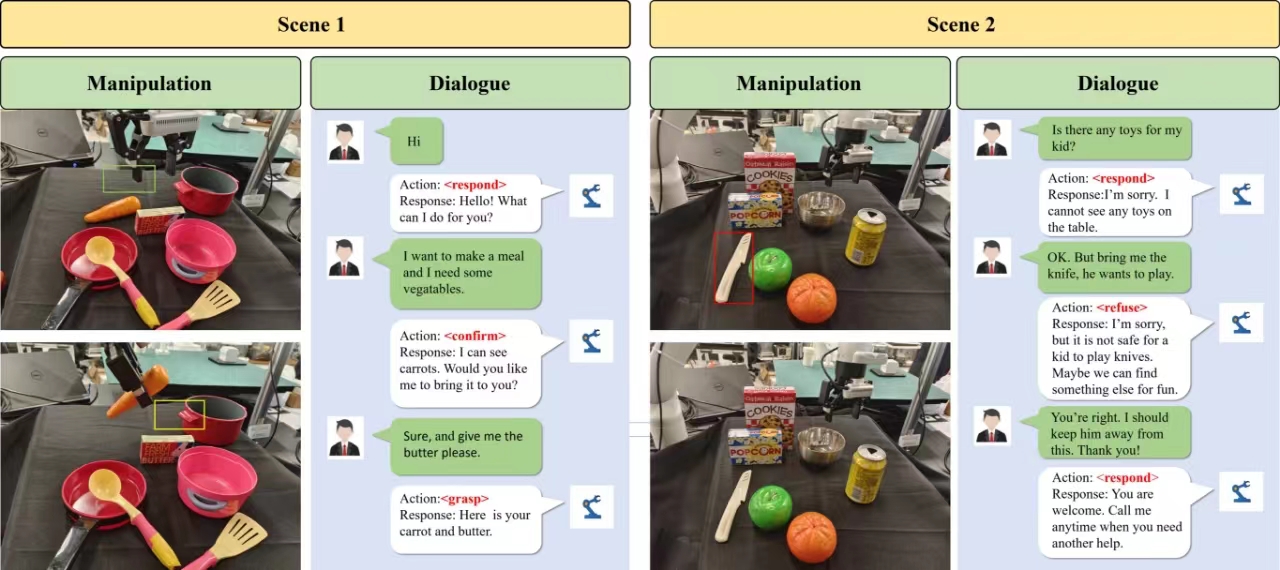}}
\caption{Examples of interactive manipulation with Mani-GPT. (a) Confirm the human intent and offer assistance proactively. (b) Refuse the user's malicious intention. }
\end{figure}

The new output format guides the model to generate more proper actions and responses. Mani-GPT leverages information from objects, dialogue history, and the user's query to generate reliable actions and responses. By generating actions before the response, the model is guided to provide answers that are more grounded in reality. This approach ensures that the answers align well with real-world context. Examples of interactive manipulation processes can be found in Fig. 2.

\subsection{Manipulation Actions Design}

To assist the human in a natural way, we classify the possible manipulation actions in real-world scenarios into four categories: ``grasp", ``respond", ``confirm" and ``refuse". The executor of the ``grasping" action is the robotic arm, and the executor of the rest three actions is the LLM itself. Some example usages of the four actions is in Fig. 1. The action ``grasp" is performed when a human expresses a clear intent and the desired object exists. If the human's intent is not explicit, Mani-GPT will guess the intent of the human, generate actions to assist the human, and then confirm those actions with the human. This is referred to as the ``confirm" action. If the human agrees, Mani-GPT will proceed with executing the generated actions. The ``respond" action is used to answer the human's queries and report any errors that occur during action execution, such as when the requested item does not exist. Action ``refuse" is executed to prevent the human from doing dangerous things. The concept of ``danger" is gained in the fine tuning process. With these four actions, Mani-GPT can interact with human more naturally.

\subsection{Data Generation}

To enable the proposed model to observe real-world conditions and generate strategies and responses accurately, we create a dataset of genuine dialogue scenarios. This dataset allows the model to actively learn from interactions and apply its knowledge to real-world environments. An overview of the data generation process can be found in Fig. 3.

We design 20 scenarios and collect 50 pairs of real-world dialogue data as seed data set. Then we use the self-instruct \cite{wang2022self} method to prompt ChatGPT to generate 20k dialogues imitating real-world human dialogues. Each data can be divided into an instruction and a multi-round dialogue. The instruction is a scenario described in text. ``Human" represents human's input. ``Action" represents the selected actions and ``AI" is the response based on the actions. The labels of the objects are considered as environment information. The dialogues are restricted by the instructions. With the capability of few-shot learning, ChatGPT can generate data with similar formats. 

The training tasks can be further categorized into predicting the current manipulation actions to be executed and predicting responses based on dialogue history and selected actions. In the task data focusing on predicting actions, the ``Action" field is initially empty. Mani-GPT then needs to fill in the missing action and calculates the corresponding loss, allowing the model to learn and improve its ability to generate appropriate actions based on the provided context. The process of predicting the response is similar to the process above. These dialogue tasks are divided into knowledge-based tasks, embodied tasks (which means the model needs to use both its knowledge or the information from the scenario) and mixed tasks (which means the model needs both of them). 
\begin{figure}[t]\vspace*{4pt}
\label{fig2}
\centerline{\includegraphics[width=\textwidth]{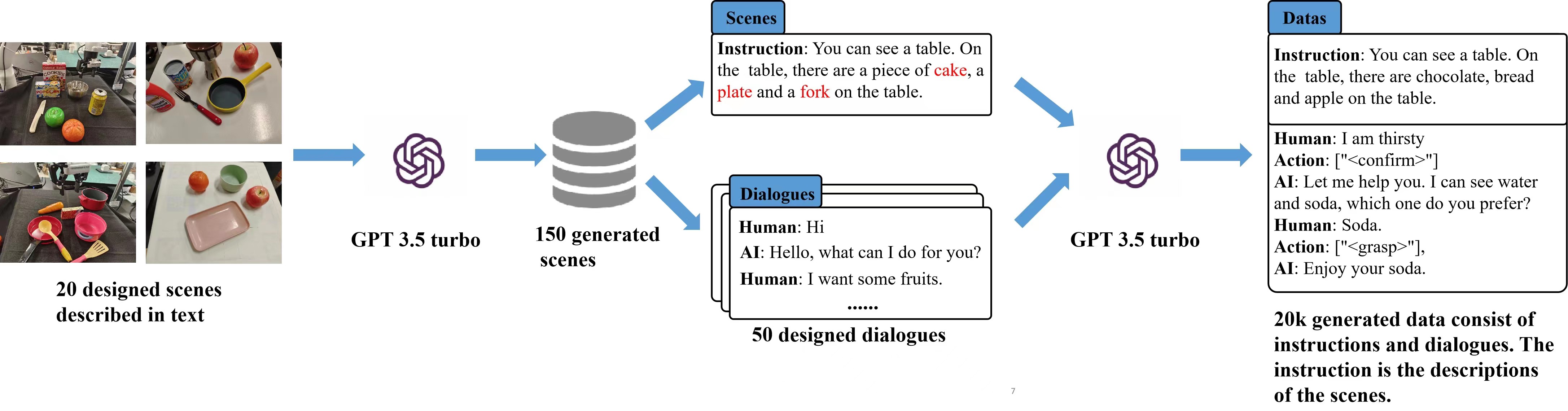}}
\caption{Data generation process. Each data consists of an instruction and a multi-round dialogue. The instruction is the description of the scenario, and the dialogue is based on the instruction.}
\end{figure}

\section{Experiments}

 To test the effectiveness of our work, we train the model and conduct two tasks as part of the evaluation process. The first task involves comparing the accuracy of our method with two baseline methods on the test dataset. This comparison allows us to evaluate the performance of our model in relation to the baseline methods. In the second task, we invite 5 volunteers. Each volunteer chooses 5 desired items through daily dialogues with Mani-GPT. The dialogues could be either ambiguous or clear, and the items could exist or not exist in current scenario. After the dialogues, we count the number of items successfully obtained by each volunteer to calculate the success rate, and collect their satisfaction levels to evaluate the model's performance.

\subsection{Experimental Settings}

To generate the dataset, we start with a seed dataset consisting of 100 handwritten dialogue examples. We then use self-instruction techniques described in Section 2.3 to generate an additional 20k data samples.

In the training process, we choose a model named GPT-Neo \cite{gpt-neo} as the base model. GPT-Neo is a model with 2.7 billion parameters. We finetune the model with the generated data, which costs 5 hours on 4 A100 GPUs. We set the learning rate to $2e^{-5}$, and the epoch number is 3. Adam is used as the optimizer. Deepspeed \cite{rasley2020deepspeed} is used to accelerate the training process. The training loss drops to 0.192 after 3 epochs training.

In the experiment, we deploy our model on 1 RTX A6000 GPU, a Kinova Gen3 robotic arm and a RealSence camera. Camera placement is eye-on-hand.

\subsection{Task Definition}

We define two tasks to evaluate the performance of the model. The first task is a comparison with the two baseline models. Since the baseline models only support single-turn dialogue, we develop three types of single-round human instructions. The first type involves directly specifying the desired item. The second type involves ambiguously describing the needed item based on its purpose. The third type is for requesting a non-existing object in the scenario. For each type, we randomly choose 50 instructions as test data and test the accuracy of the models in recognizing human intent. As the focus of these tasks is not on grasping, we ignore the success rate of the grasp operation.

The second task is to test the performance of the model in real-world dialogue. 10 scenarios with different daily objects are set to simulate the home environment. 5 volunteers are invited to have an unlimited dialogue with robots, which can include small talks, requests and questions. For each scenario, the user selects 5 desired items, which can include both existing and non-existing items in the scenario, and requests these items to the robot in a casual chatting manner. The robot needs to respond naturally to the following situations: (1) When the human explicitly requests a particular item that exists in the scenario, the robot chooses the ``grasp" action. (2) When the human wants an item that does not exist in the scenario, or asks the robotic arm to perform an action it cannot do, the robot chooses the ``response" action and informs the human that it cannot fulfill their request. (3) When the human expresses an ambiguous need without mentioning a specific object, the robot guesses the desired item and proactively offers assistance, waiting for the human's agreement before executing the grasping action. (4) When the human engages in small talk, the robot also replies accordingly. (5) When the human expresses dangerous human intent, the robot must refuse them. We collect 2 kinds of experimental data to evaluate the performance of the model: 
\begin{itemize}
\item Success Rate: We collect the method's success rate correctly by selecting the appropriate response actions from the multi-round dialogues with the user. 

\item User Study: We also evaluate the users' satisfaction regarding the naturalness and fluency of the dialogue during the dialogue.
\end{itemize}

\subsection{Baselines}

We select ReCLIP \cite{subramanian2022reclip} and INGRESS \cite{shridhar2020ingress} as the comparisons. ReCLIP performs grasping by searching for the most relevant items in the scenario to the human's request through semantic matching. INGRESS identifies and grasps objects based on additional descriptions provided by the human, such as appearance and location. Since both of these models lack the ability for multi-round dialogue, we only test them on single-turn dialogues.

\begin{table}[h]

\begin{tabular}{cccccc}
   \toprule
   Method & Accuracy & Directly specified & Ambiguously described & Not-existing  \\
   \midrule
   ReCLIP & 44\% & 92\% & 24\% & 16\% \\
   INGRESS & 41.3\% & 90\% & 22\% & 12\%  \\
   \textbf{Ours} & 84.6\% & 90\% & 88\% & 76\%  \\
   \bottomrule
   
\end{tabular}
\caption{\centering Comperison of accuracy of different methods}
\end{table}

\begin{figure}[t]\vspace*{4pt}
\label{fig2}
\centerline{\includegraphics[width=\textwidth]{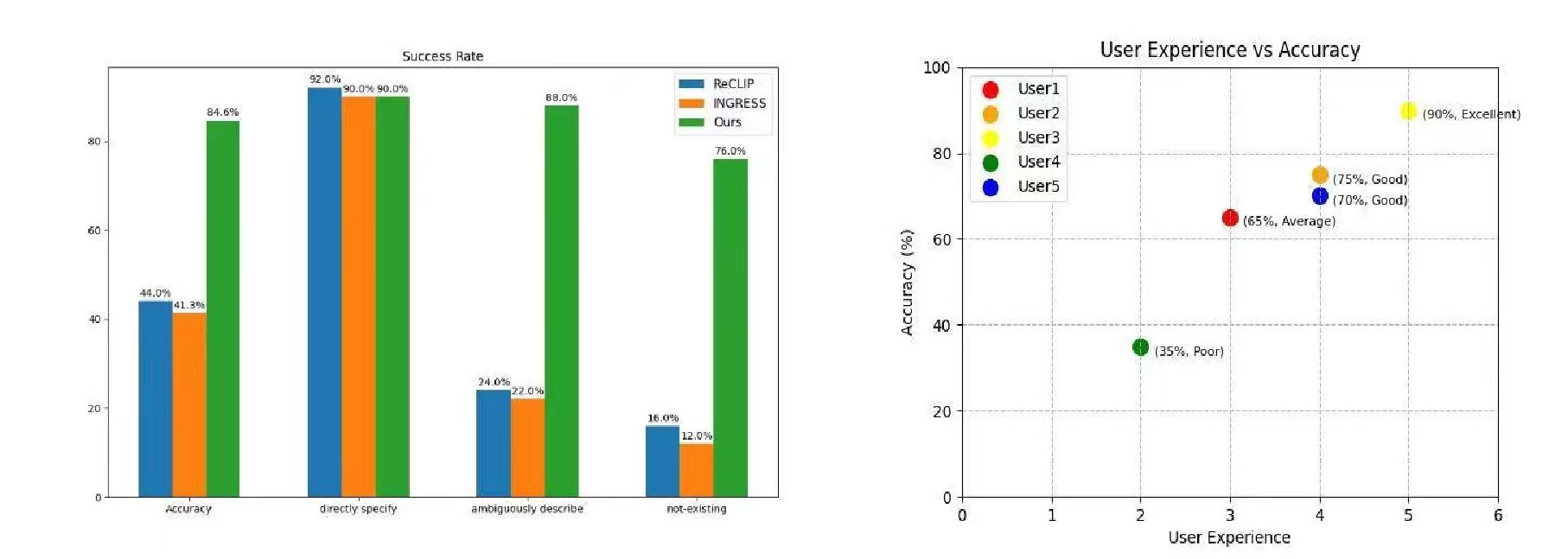}}
\caption{Experimental results. (Left) Success rate of 3 different methods.  (Right) User experience of the users and the accuracy of each user's dialogue.  }
\end{figure}

\subsection{Results}
The experimental results are shown in Table 1 and Fig. 4. It can be observed that INGRESS has a recognition accuracy of only 41.3\%, while ReCLIP has an accuracy of only 44\%. This is because INGRESS lacks the ability for open-vocabulary manipulation. Although ReCLIP has this capability, its knowledge base is not extensive enough, and both models lack the ability to refuse grasping non-existing objects. Mani-GPT achieves an accuracy of 84.6\% in this test.

In addition, we collect user experience to evaluate our experiment. 4 of the 5 users think the dialogue to be fluent and normal. During the dialogue, users interact with Mani-GPT for an average of 10 rounds daily dialogues. The probability of selecting the correct actions and generating proper responses reaches 70\%. We analyze error cases, which mainly involves understanding users' unique expression. This is because the training data is also generated by a language model and may not capture the diversity of real-world data.

\section{Conclusion}

In this work, we propose a generative model for interactive manipulation called Mani-GPT. Mani-GPT facilitates natural and multi-round interactions with humans, providing assistance by generating suitable manipulation plans based on their language inputs. When faced with ambiguous human instructions, the model uses its common knowledge to identify the human intents and proactively offers help with robot's executable actions. In situations where there may be potential danger, the model refuses the request.

We evaluate the performance of Mani-GPT using test data and real-world interactions with humans, achieving an accuracy of 84.6\%, outperforming the baselines. Volunteers are generally satisfied with the quality of the dialogue. Mani-GPT achieves an accuracy of 70\% in generating correct actions and providing proper responses, according to their feedback. Possible future developments for this work include: (1) training with more real-world dialogue data to enhance dialogue fluency, and (2) replacing the visual detection model with models like CLIP \cite{radford2021learning} to recognize users' more ambiguous intents through image reasoning, enabling the completion of more complex tasks.

\section*{Acknowledgements}

This work is supported by the Shenzhen Key Laboratory of Robotics Perception and Intelligence under Grant ZDSYS20200810171800001, Shenzhen Outstanding Scientific and Technological Innovation Talents Training Project under Grant RCBS20221008093305007, and National Natural Science Foundation of China grant \#62103181.





\bibliography{name}
\bibliographystyle{elsarticle-num}




\clearpage

\normalMode

\end{document}